\documentclass[lettersize,journal]{IEEEtran}
\usepackage{amsmath,amsfonts}
\usepackage{algorithmic}
\usepackage{array}
\usepackage[caption=false,font=normalsize,labelfont=sf,textfont=sf]{subfig}
\usepackage{textcomp}
\usepackage{stfloats}
\usepackage{url}
\usepackage{verbatim}
\usepackage{graphicx}
\usepackage{cite}

\usepackage{xcolor,soul,framed} 
\colorlet{shadecolor}{yellow}
\graphicspath{{./pdf/}{./jpeg/}}
\DeclareGraphicsExtensions{.pdf,.jpeg,.png}
\usepackage{array}
\usepackage{booktabs}
\usepackage{mdwmath}
\usepackage{mdwtab}
\usepackage{eqparbox}
\usepackage{url}
\usepackage{pifont}
\usepackage{amssymb}
\usepackage{multirow}
\usepackage{diagbox}
\usepackage{makecell}
\usepackage[colorlinks,
            linkcolor=blue,
            anchorcolor=blue,
            citecolor=red]{hyperref}
\usepackage{threeparttable}
\usepackage[ruled]{algorithm2e}
\usepackage{float}

\begin{document}

\title{Dual Radar: A Multi-modal Dataset with Dual 4D Radar for Autononous Driving}

\author{
Xinyu Zhang$^*$, Li Wang$^*$, Jian Chen, Cheng Fang, Lei Yang, Ziying Song, Guangqi Yang, Yichen Wang, Xiaofei Zhang, Jun Li,  Zhiwei Li, Qingshan Yang, Zhenlin Zhang, Shuzhi Sam Ge

\thanks{This work was supported by the National High Technology Research and Development Program of China under Grant No. 2018YFE0204300, the National Natural Science Foundation of China under Grant No. 62273198, U1964203, 52221005. \emph{(Corresponding author: Li Wang. *: These authors contributed equally to this work.)}}

\thanks{Xinyu Zhang, Li Wang, Lei Yang, Ziying Song, Guangqi Yang, Yichen Wang, Xiaofei Zhang, and Jun Li are with the State Key Laboratory of Automotive Safety and Energy, and the School of Vehicle and Mobility, Tsinghua University, Beijing 100084, China (e-mail: 
xyzhang@tsinghua.edu.cn;
lwang\_hit@hotmail.com;
yanglei20@mails.tsinghua.edu.cn;
22110110@bjtu. edu.cn;
yanggq23@mails.jlu.edu.cn;
22S136042@stu.hit.edu.cn;
csezxf@ 126.com;
lijun19580326@126.com).
}

\thanks{Jian Chen is with the School of Mechanical and Electrical Engineering, China University of Mining and Technology-Beijing, Beijing 100083, China (e-mail: ZQT2100407181@student.cumtb.edu.cn).
}
\thanks{Cheng Fang is with the School of Artificial Intelligence, China University of Mining and Technology-Beijing, Beijing 100083, China (e-mail: sqt2100407123@student.cumtb.edu.cn).
}




\thanks{Zhiwei Li is with the College of Information Science and Technology, Beijing University of Chemical Technology, Beijing 100029, China (e-mail: 2022500066@buct.edu.cn).
}

\thanks{Qingshan Yang is with Beijing Jingwei Hirain Technologies Co., Inc., Beijing 100191, China (e-mail: qingshan.yang@hirain.com).
}

\thanks{Zhenlin Zhang is with China Automotive Innovation Cooperation, Nanjing 211113, China (e-mail: Zhenlin.zhang@gmail.com).
}

\thanks{Shuzhi Sam Ge is with the Department of Electrical and Computer Engineering, Interactive Digital Media Institute, Social Robotics Laboratory, National University of Singapore, Singapore (e-mail: samge@nus.edu.sg).
}
}

\maketitle

\begin{abstract}
Radar has stronger adaptability in adverse scenarios for autonomous driving environmental perception compared to widely adopted cameras and LiDARs. Compared with commonly used 3D radars, latest 4D radars have precise vertical resolution and higher point cloud density, making it a highly promising sensor for autonomous driving in complex environmental perception. However, due to the much higher noise than LiDAR, manufacturers choose different filtering strategies, resulting in a direct ratio between point cloud density and noise level. There is still a lack of comparative analysis on which method is beneficial for deep learning-based perception algorithms in autonomous driving. One of the main reasons is that current datasets only adopt one type of 4D radar, making it difficult to compare different 4D radars in the same scene. Therefore, in this paper, we introduce a novel large-scale multi-modal dataset featuring, for the first time, two types of 4D radars captured simultaneously. This dataset enables further research into effective 4D radar perception algorithms. 
Our dataset consists of 151 consecutive series, most of which last 20 seconds and contain 10,007 meticulously synchronized and annotated frames. Moreover, our dataset captures a variety of challenging driving scenarios, including many road conditions, weather conditions, nighttime and daytime with different lighting intensities and periods. Our dataset annotates consecutive frames, which can be applied to 3D object detection and tracking, and also supports the study of multi-modal tasks. We experimentally validate our dataset, providing valuable results for studying different types of 4D radars. This dataset is released on https://github.com/adept-thu/Dual-Radar.
\end{abstract}



\begin{figure}
  \begin{center}
  \includegraphics[width=3.2in]{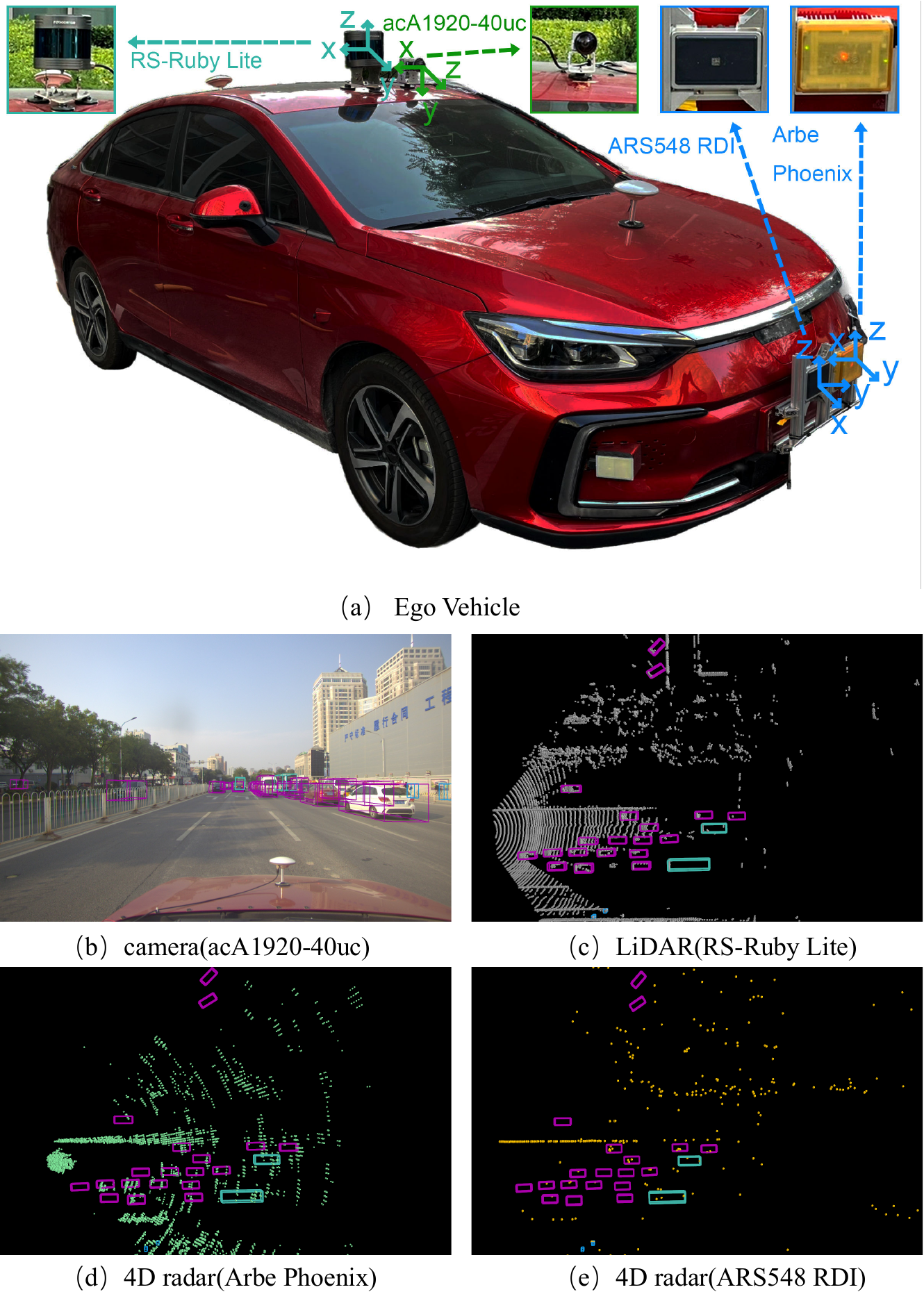} \\
  \caption{The configuration of our experiment platform and visualization scenarios on the data collected by different sensors. (a) shows information of each sensor coordinate system in the self-driving car system. (b), (c), (d), and (e) show the results of the 3D bounding box annotations on the data (images, LiDAR point cloud, Arbe Phoenix point cloud, ARS548 RDI point cloud).}\label{circuit_diagram}
  \end{center}
\end{figure}

\section{Introduction}

\IEEEPARstart{A}{s} the key aspect of autonomous driving technology, environmental perception can timely detect the external things that affect the safety of driving in the process and provide the basis for the subsequent decision-making and control links, which guarantee the safety and intelligence of driving\cite{wang2020ethical,wang2019cyber,huang2019novel,wen2020edge,graphalign++}. In recent years, sensors such as cameras, LiDAR, and radars in ego vehicles are attracting research interest due to remarkable increases in the performance of sensors and the computers’ arithmetic power\cite{arnold2019survey,herzog2019training,danzer20192d,wang2023interference,survey,VP-net}.

\begin{table*}[h!]
  \begin{center}
    \renewcommand\arraystretch{1.1}
    \centering
    \caption{Comparison of configurations of 4D radar in commonly datasets}
    \begin{tabular}{ccccccccccc}
    \toprule[1pt]
        \multirow{2}{*}{{\textbf{Dataset}}} & \multirow{2}{*}{{\textbf{Radar Type}}}    & \multicolumn{3}{c}{{\textbf{Resolution}}}    & & \multicolumn{3}{c}{{\textbf{FOV}}}                   & \multirow{2}{*}{{\textbf{FPS}}} & \multirow{2}{*}{{\textbf{Radar mode}}} \\
        \cline{3-5} \cline{7-9}
                         &                          & {\textbf{Range}} & {\textbf{Azimuth}} & {\textbf{Elevation}}  &   &
                         {\textbf{Range}}            & {\textbf{Azimuth}}       & {\textbf{Elevation}}     &                      &                             \\
    \midrule[0.4pt]
        Astys\cite{meyer2019automotive}                    & \makecell[c]{Astyx 6455 \\ HiRes}              & -  &  -            & -             &     & 100m  & 110°    & 10°         & 13                   & Middle                      \\
        RADIal\cite{rebut2022raw}                   & -                    & 0.2m             & -         & -   &    & 103 m & 180°    & 12°         & 5                    & Middle                      \\
        View-of-Delft\cite{palffy2022multi}            & \makecell[c]{ZF FRGen 21 \\  3+1D Radar} & \textless{}=0.2m & 1.5°          & 1.5°  &   & -     & -       & -              & 13                   & Short                       \\
        TJ4DRaDSet\cite{zheng2022tj4dradset}               & -                & 0.86m            & \textless{}1° & \textless{}1° &  & 400m  & 113°    & 45°       & 15                   & Long                       \\
        K-Radar\cite{paek2022k}                  & RETINA-4ST         & 0.46m            & 1°            & 1°   & & 118m  & 107°    & 37°         & 10                   & Middle    \\           

        \multirow{2}{*}{Ours} & ARS548 RDI & 0.22m & \makecell[c]{1.2\textdegree @0…$\pm$15\textdegree\\ 1.68\textdegree @ $\pm$45\textdegree }& 2.3\textdegree & & 300m & $\pm$60\textdegree  &  \makecell[c]{$\pm$4\textdegree@300m\\ $\pm$14\textdegree@$<$100m }& 20 &Long \\

         & Arbe Phoenix & 0.3m & 1.25 \textdegree & 2\textdegree & & 153.6m & 100\textdegree & 30\textdegree & 20 &Middle \\
    \bottomrule[1pt]
    \end{tabular}

    \begin{tablenotes}{    
          \footnotesize           
          \item[1] “Short”: short-range mode. “Middle”: middle-range mode. “Long”: long-range mode.
          }
    \end{tablenotes}
  \label{tab:table1}
\end{center}
\end{table*}


\begin{table*}[h!]
  \begin{center}
  \setlength{\tabcolsep}{11.3pt}
  \renewcommand\arraystretch{1.1}
  \centering
  \caption{the overview of the configuration of the datasets for object detection and tracking with radar mentioned}
  
      \begin{tabular}{cccccccc}
          \toprule[1pt]
            \multirow{2}{*}{{\textbf{Dataset}}} & \multirow{2}{*}{{\textbf{Year}}} & \multirow{2}{*}{{\textbf{Size}}} & {{\textbf{Synchronous}}}&\multicolumn{2}{c}{{\textbf{Tasks}}} & \multirow{2}{*}{{\textbf{Doppler}}} & \multirow{2}{*}{{\textbf{Radar mode}}}  \\
            \cline{5-6} & & &{\textbf{(frames)}} & {\textbf{3D Detection}} & {\textbf{3D Tracking}} & &   \\
          \midrule[0.4pt]
          \midrule[0.4pt]
          \multicolumn{8}{c}{{\textbf{\uppercase{the dataset with 3D radar mentioned}}}} \\
          \midrule[0.4pt]
          \midrule[0.4pt]
          NuScenes\cite{caesar2020nuscenes} & 2020 & Large & 400K & $\checkmark$ & $\checkmark$  & \ding{55} & Long,Surrounding  \\
          Pointillism\cite{bansal2020pointillism} & 2020 & Medium & 54K & $\checkmark$ & \ding{55}  & $\checkmark$ & Middle  \\
          Zendar\cite{mostajabi2020high} & 2020 & Small & 4,780 &$\checkmark$ & \ding{55}  & $\checkmark$ & Middle  \\
          Dense\cite{bijelic2020seeing} & 2020 & Large & 13.5K & $\checkmark$ & \ding{55} & $\checkmark$ & Long  \\
          RADIATE\cite{sheeny2021radiate} & 2020 & Medium & 44K & $\checkmark$ & $\checkmark$ & \ding{55} & Middle,Surrounding  \\      
          PixSet\cite{deziel2021pixset} & 2021 & Medium & 29K & $\checkmark$ & $\checkmark$ & $\checkmark$ & Middle   \\
          RadarScence\cite{schumann2021radarscenes} & 2021 & Large & 832K & $\checkmark$ & $\checkmark$ & $\checkmark$ & Middle,Surrounding  \\
          \midrule[0.4pt]
          \midrule[0.4pt]
          \multicolumn{8}{c}{{\textbf{\uppercase{the dataset with 4D radar mentioned}}}} \\
          \midrule[0.4pt]
          \midrule[0.4pt]
          Astyx\cite{meyer2019automotive} & 2019 & Small & 546 & $\checkmark$ & \ding{55} & $\checkmark$ & Middle  \\
          RADIal\cite{rebut2022raw} & 2021 & Large & 25K & $\checkmark$ & \ding{55} & $\checkmark$ & Middle  \\
          View-of-Delft\cite{palffy2022multi} & 2022 & Medium & 8,693 & $\checkmark$ & $\checkmark$ & $\checkmark$ & Short  \\
          TJ4DRaDSet\cite{zheng2022tj4dradset} & 2022 & Medium & 7,757 & $\checkmark$ & $\checkmark$  & $\checkmark$ & Long  \\
          K-Radar\cite{paek2022k} & 2022 & Large & 35K & $\checkmark$ & $\checkmark$ &  $\checkmark$ & Middle  \\
          Ours & 2023 & Large & 10K & $\checkmark$ & $\checkmark$ & $\checkmark$ & Long\&Middle  \\
          \bottomrule[1pt]
      \end{tabular}
        
        \begin{tablenotes}{    
          \footnotesize           
          \item[1] “Short”: short-range mode. “Middle”: middle-range mode. “Long”: long-range mode. “Surrounding”: collection data within a 360° range.
          }
    \end{tablenotes}
    
    \label{tab:overview3D}
  \end{center}
\end{table*}


\begin{table*}[h!]
  \begin{center}
  \renewcommand\arraystretch{1.1}
  \setlength{\tabcolsep}{9.5pt}
  \caption{the overview of the details of the datasets for object detection and tracking with radar mentioned}
  \centering
      \begin{tabular}{ccccccccccc}
          \toprule[1pt]
            \multirow{2}{*}{{\textbf{Dataset}}} & \multirow{2}{*}{{\textbf{Year}}} & \multicolumn{5}{c}{{\textbf{Scenarios}}} & \multirow{2}{*}{{\textbf{Weather}}} & \multicolumn{3}{c}{{\textbf{Annotations}}}  \\
            \cline{3-7, 9-10} & & {\textbf{Urban}} & {\textbf{Suburban}} & {\textbf{Highway}} & {\textbf{Tunnel}} & {\textbf{Parking}} & & {\textbf{3D BOX}} & {\textbf{Track ID}} & \\
            
          \midrule[0.4pt]
          \midrule[0.4pt]
          \multicolumn{10}{c}{{\textbf{\uppercase{the dataset with 3D radar mentioned}}}} \\
          \midrule[0.4pt]
          \midrule[0.4pt]
          nuScense\cite{caesar2020nuscenes} & 2020 & $\checkmark$ & $\checkmark$ & \ding{55} & $\checkmark$ & \ding{55} & $\checkmark$ & $\checkmark$ & $\checkmark$ \\
          Pointillism\cite{bansal2020pointillism} & 2020 & $\checkmark$ & \ding{55} & \ding{55} & \ding{55} & \ding{55} & $\checkmark$ & $\checkmark$ & \ding{55}\\
          Zendar\cite{mostajabi2020high} & 2020 & $\checkmark$ & \ding{55} & \ding{55} & \ding{55} & \ding{55} & \ding{55} & \ding{55} & \ding{55} \\
          Dense\cite{bijelic2020seeing} & 2020 & $\checkmark$ & $\checkmark$ & $\checkmark$ & $\checkmark$ & \ding{55} & $\checkmark$ & $\checkmark$ & \ding{55}\\
          RADIATE\cite{sheeny2021radiate} & 2020 & $\checkmark$ & $\checkmark$ & $\checkmark$ & \ding{55} & $\checkmark$ & $\checkmark$ & \ding{55} & $\checkmark$\\
          PixSet\cite{deziel2021pixset} & 2021 & $\checkmark$ & $\checkmark$ & \ding{55} & \ding{55} & $\checkmark$ & $\checkmark$ & $\checkmark$ & $\checkmark$\\
          RadarScence\cite{schumann2021radarscenes} & 2021 & $\checkmark$ & $\checkmark$ & $\checkmark$ & $\checkmark$ & \ding{55} & $\checkmark$ & \ding{55} & \ding{55} \\
          \midrule[0.4pt]
          \midrule[0.4pt]
          \multicolumn{10}{c}{{\textbf{\uppercase{the dataset with 4D radar mentioned}}}} \\
          \midrule[0.4pt]
          \midrule[0.4pt]
          Astyx\cite{meyer2019automotive} & 2019 & \ding{55} & $\checkmark$ & $\checkmark$ & \ding{55} & \ding{55} & \ding{55} & $\checkmark$ & \ding{55}\\
          RADIal\cite{rebut2022raw} & 2021 & $\checkmark$ & $\checkmark$ & $\checkmark$ & \ding{55} & \ding{55} & \ding{55} & \ding{55} & \ding{55} \\
          View-of-Delft\cite{palffy2022multi} & 2022 & $\checkmark$ & \ding{55} & \ding{55} & \ding{55} & \ding{55} & \ding{55} & $\checkmark$ & $\checkmark$ \\
          TJ4DRaDSet\cite{zheng2022tj4dradset} & 2022 & $\checkmark$ & \ding{55} & \ding{55} & \ding{55} & \ding{55} & \ding{55} & $\checkmark$ & $\checkmark$ \\
          K-Radar\cite{paek2022k} & 2022 & $\checkmark$ & $\checkmark$ & $\checkmark$ & $\checkmark$ & \ding{55} & $\checkmark$ & $\checkmark$ & $\checkmark$ \\
          Ours & 2023 & $\checkmark$ & \ding{55} & \ding{55} & $\checkmark$ & \ding{55} & $\checkmark$ & $\checkmark$ & $\checkmark$ \\
          \bottomrule[1pt]
      \end{tabular}
    
     \begin{tablenotes}{    
          \footnotesize           
          \item[1] 3D BOX: 3D Bounding Box. Track ID: Tracking ID.
          }
    \end{tablenotes}
    \label{tab:overview4D}
\end{center}
\end{table*}

The camera has high resolution, enabling true RGB information and abundant semantic features such as colors, categories, and shapes. However, when meeting quickly changeable or weak-intensity light, the results obtained from the environmental perception task using only the camera are undesirable\cite{chen2016monocular,chen20153d,mousavian20173d,Bevheight,Mixteaching}. Moreover, monocular cameras cannot accurately acquire distance information, and multi-camera or fisheye cameras suffer from obvious lens distortion problems\cite{secci2020failures, wu2022integrated}. LiDAR can collect dense 3D point clouds, with the advantages of high density and accurate precision\cite{tan20223d,wang2023camo}. When the autonomous vehicle is driving at high speed, the LiDAR still operates in a mechanical full-view rotation to collect data, and this driving condition can lead to distortion of the point cloud obtained by the LiDAR\cite{rashed2019fusemodnet, xue2019real}. Meanwhile, LiDAR performs poorly in adverse weather due to its reliance on optical signals to acquire point clouds\cite{yoneda2019automated, zang2019impact, vargas2021overview}. Radar, on the other hand, opens up new prospects in autonomous vehicles due to its recognized advantages such as small size, low cost, all-weather operation, high-speed measurement capability, and high range resolution\cite{xu2021rpfa,abdu2021application,zheng2023rcfusion}. In particular, radar works with electromagnetic wave signals, has good penetration performance in adverse weather, and has a long propagation distance, which makes up for the shortcomings of cameras and LiDAR\cite{jiang20234d,mostajabi2020high,sheeny2021radiate}. Currently, the popular radar sensors in automated driving mainly include 3D  radar and 4D radar. 4D radar can provide three-dimensional information, including distance, azimuth, and elevation, as well as a denser point cloud and more elevation information than 3D radar, which is beneficial for dealing with driving scenarios in adverse conditions\cite{stolz2018new, sun20214d, zhou2022towards}.

Due to the sparsity, 4D radar collects less information than LiDAR point clouds. Although 4D radar performs well in unfavorable scenarios, its sparsity can still lead to the possibility of missing objects. With the development of radar sensors, various types of 4D radar have been applied in scientific research\cite{wang2022multi}. 4D radar has different working modes according to the detection range, which can be classified into short-range, middle-range, and long-range modes\cite{kramer2022coloradar}. In Table \ref{tab:table1}, each 4D radar can be observed to have its advantages. In different working modes, the resolution and range of point clouds collected by 4D radars are different, and 4D radars have different point cloud densities and collection ranges in multiple working modes\cite{kramer2022coloradar}. As shown in Fig. \ref{circuit_diagram}(d), due to the large beamwidth of Arbe Phoenix, which usually does not process noise, the point cloud is dense, but it will lead to the possibility of false detection. On the other hand, as shown in Fig. \ref{circuit_diagram}(e), the point cloud collected by the ARS548 RDI sensor is sparse after noise processing, which may lead to missing objects and is not favorable for the detection of small or close proximity objects. However, the denoising process can improve the object detection accuracy of 4D radar. At the same time, the ARS548 RDI can collect a long-range of point clouds than the Arbe Phoenix, which means that the ARS548 RDI can collect information about objects at longer distances. In the field of autonomous driving, existing datasets are usually studied using only one type of 4D radar, lacking comparative analysis of 4D radars with different point cloud densities and levels of noise in the same scenario, as well as a lack of research on perception algorithms that can process different types of 4D radars.

To validate the performance of different types of 4D radar in object detection and object tracking tasks and to fulfill researchers' needs for a 4D radar dataset, we propose a novel dataset with two types of 4D radar point clouds. Our proposed dataset is collected from a high-resolution camera, an 80-line mechanical LiDAR, and two types of 4D radars, the Arbe Phoenix and the ARS548 RDI radar. Our dataset provides GPS information for timing implementation. The sensor configurations are shown in Table \ref{tab:configuration}. Collecting two types of 4D radar point clouds can explore the performance of point clouds with different sparsity levels for object detection in the same scenario, which will provide a basis for developing 4D radar research in the field of ego vehicles.


Our main contributions to this work are as follows:

(1) We present a dataset with multi-modal data, which includes camera data, LiDAR point cloud, and two types of 4D radar point cloud. Our dataset can study the performance of different types of 4D radar data, contributes to the study of perception algorithms that can process different types of 4D radar data, and can be used to study single-modal and multi-modal fusion tasks.

(2) Our dataset provides a variety of challenging scenarios, including scenarios with different road conditions (city and tunnel), different weather (sunny, cloudy, and rainy), different light intensities (normal light and backlight), different times of day (daytime, dusk, and nighttime), which can be used to study the performance of different types of 4D radar point cloud in different scenarios.

(3) Our dataset consists of 151 continuous time sequences, with most time sequences lasting 20 seconds. There are 10,007 frames carefully synchronized in time and 103,272 high-quality annotated objects.



The main organization of this paper is as follows.

In Section \ref{sec2:relatedwork}, we presented related work on using radar for object detection and tracking datasets. Section \ref{sec3:dataset} provided specific information about the dataset, including its main content and visualizations of different scenarios. The experimental details are described in Section \ref{sec4:experiment}. Finally, in Section \ref{sec5:conclusion}, we summarized our work, discussed some limitations of the current study, and outlined future research directions.


\section {Related Work} \label{sec2:relatedwork}

Many companies and institutions have recently proposed their own datasets for research on autonomous vehicle systems based on multi-modal fusion, as shown in Table \ref{tab:overview3D} and Table \ref{tab:overview4D}. Early datasets primarily utilized 3D radar to collect data from the environment within a horizontal field of view. However, the emergence of 4D radar sensors fills the limitations of 3D radar in challenging scenarios, providing elevation information and enabling researchers to explore more meaningful research topics in the realm of 4D radar.


\subsection {The Dataset with 3D radar}

Numerous datasets and applications based on traditional 3D radar have been released\cite{bijelic2020seeing, caesar2020nuscenes, deziel2021pixset, schumann2021radarscenes, bansal2020pointillism, mostajabi2020high, sheeny2021radiate}. These datasets provide important references regarding dataset scale, sensor types, and data processing methods. One early dataset related to 3D radar is nuScenes\cite{caesar2020nuscenes}, which has a large-scale with rich scenarios and object categories. It contains 140,000 object labels in 40,000 annotated frames, including scenarios with adverse weather and varying lighting conditions. This dataset fully leverages the advantages of radar, providing abundant information, such as 3D bounding boxes and tracking IDs, for object detection and tracking tasks in challenging scenarios and supporting localization and mapping tasks. Using this dataset, researchers have produced promising results in object detection and tracking. However, the low resolution of the 3D radar in nuScenes does not meet current research demands. Pointillism\cite{bansal2020pointillism} offers a dataset of medium-scale with adverse weather conditions. It provides data for 3D radar in a middle-range mode, with object detection range from 120m to 150m. However, Pointillism only provides one scenario and does not offer tracking IDs for object tracking tasks. Zendar\cite{mostajabi2020high} has a rich variety of data types, but the dataset scale is relatively small, with only around 50,000 annotated objects in 4,780 frames. Moreover, it has a single weather condition and does not support object tracking research. Dense\cite{bijelic2020seeing} is a large-scale dataset with a radar data collection range of less than 230m. It contains various scenarios, including adverse weather conditions like heavy fog, snow, and rain during daylight and night. However, this dataset focuses on object detection tasks across multiple scenarios and does not provide tracking IDs for object tracking research. RADIATE\cite{sheeny2021radiate} has a dataset of medium-scale with diverse driving scenarios and weather conditions applicable for localization and mapping tasks. However, the data type is not point cloud, making it inappropriate for feature-level fusion research between LiDAR and radar in object detection tasks. The PixSet\cite{deziel2021pixset} dataset has a medium-scale size. Compared to nuScenes, PixSet has more road scenarios and adverse weather conditions but smaller data.  RadarScene\cite{schumann2021radarscenes} provides massive data through high-resolution 3D radar. It contains a variety of object categories, driving scenarios, and adverse weather conditions, making it applicable for object detection and localization tasks. However, the point cloud in RadarScene is annotated at the point level and does not have 3D bounding boxes.

\subsection {The Dataset with 4D radar}

Since the release and application of 4D radar, several datasets with 4D radar have been published\cite{meyer2019automotive, palffy2022multi, rebut2022raw, zheng2022tj4dradset, paek2022k}. Astyx\cite{meyer2019automotive}, an early-released 4D radar dataset, provides rich data for 3D object detection and has been the subject of numerous academic studies in recent years. However, the total count for Astyx is close to the small size of 546 frames and contains only about 3000 object annotations. In addition, the Astyx dataset has no special scenarios, and the lack of urban data in the dataset does not meet the needs of the current research. RADIal\cite{rebut2022raw} offers a medium-scale dataset with urban streets, highways, and rural roads. The dataset is suitable for object detection tasks, but it lacks 3D bounding boxes for 3D object detection and tracking IDs for object tracking research. Furthermore, it does not capture scenarios under adverse weather conditions, limiting its use in studying challenging scenarios. View-of-Delft\cite{palffy2022multi} is a recently released dataset that better addresses problems related to object tracking. It comprises 8,693 frames with 120,000 annotated objects, making it a medium-scale dataset suitable for advancing object detection research. However, View-of-Delft has a short range for object detection and lacks 4D radar information in long-range mode. TJ4DRaDSet\cite{zheng2022tj4dradset} has a large amount of data, including various driving scenarios such as elevated roads, complex intersections, and city roads. It includes scenarios under different lighting conditions, effectively supporting object detection and tracking tasks. However, this dataset collects 4D radar point clouds in long-range mode, while data collection in middle-range and short-range modes is missing. Additionally, TJ4DRaDSet does not include scenarios with adverse weather conditions. K-Radar\cite{paek2022k} provides rich driving scenarios with various roads (city, suburban roads, alleys, and highways), adverse weather (fog, rain, and snow), and different periods ( daylight and night), with a total of 35K frames of data and 93K annotated objects. However, the K-Radar dataset lacks the 4D radar point cloud in long-range mode.


\section {Dual Radar Dataset}\label{sec3:dataset}

In this section, we propose the main aspects of our dataset, including the sensor specification of the ego vehicle system, sensor calibration, dataset annotation, data collection and distribution, and the visualization of the dataset.

\subsection {Sensor Specification}

Our ego vehicle's configuration and the coordinate relationships between multiple sensors are shown in Fig. \ref{circuit_diagram}. The platform of our ego vehicle system consists of a high-resolution camera, a new 80-line LiDAR, and two types of 4D radar. All sensors have been carefully calibrated. The camera and LiDAR are mounted directly above the ego vehicle, while the 4D radars are installed in front of it. Due to the range of horizontal view limitations of the camera and 4D radars, we only collect data from the front of our ego vehicle for annotation. The ARS548 RDI captures data within approximately 120° horizontal field of view and 28° vertical field of view in front of the ego vehicle, while the Arbe Phoenix, operating in middle-range mode, collects data within a 100° horizontal field of view and 30° vertical field of view. The LiDAR collects around the ego vehicle in a 360° manner but only retains the data in the approximate 120° field of view in front of it for annotation. The specifications of the sensors are provided in Table \ref{tab:configuration}.

\begin{table*}[h!]
    \begin{center}
    \renewcommand\arraystretch{1.1}
    \centering
    \setlength{\tabcolsep}{10pt}
    \caption{the configuration of the autonomous vehicle system platform}
    \begin{tabular}{cccccccccc}
        \toprule[1pt]
            \multirow{2}{*}{{\textbf{Sensors}}}& \multirow{2}{*}{{\textbf{Type}}} & \multicolumn{3}{c}{{\textbf{Resolution}}} & & \multicolumn{3}{c}{{\textbf{FOV}}} & \multirow{2}{*}{{\textbf{FPS}}}  \\
            \cline{3-5,7-9} & & {\textbf{Range}} & {\textbf{Azimuth}} & {\textbf{Elevation}} & & {\textbf{Range}} & {\textbf{Azimuth}} & {\textbf{Elevation}} &  \\
        \midrule[0.4pt]
        Camera & acA1920-40uc & -- & 1920px & 1200px & & -- & -- & -- & 10 \\
        LiDAR & RS-Ruby Lite & 0.05m & 0.2\textdegree & 0.2\textdegree & & 230m & 360\textdegree & 40\textdegree & 10 \\
        \multirow{2}{*}{4D radar} & ARS548 RDI & 0.22m &   \makecell[c]{1.2\textdegree @0…$\pm$15\textdegree\\ 1.68\textdegree @ $\pm$45\textdegree} & 2.3\textdegree & & 300m & $\pm$60\textdegree &  \makecell[c]{ $\pm$4\textdegree@300m\\ $\pm$14\textdegree@$<$100m }& 20 \\
         & Arbe Phoenix & 0.3m & 1.25 \textdegree & 2\textdegree & & 153.6m & 100\textdegree & 30\textdegree & 20 \\
        \bottomrule[1pt]
    \end{tabular}
    
  \label{tab:configuration}
\end{center}
\end{table*}

\subsection {Sensor Calibration}
\begin{figure}
  \begin{center}
  \includegraphics[width=3.3in]
  {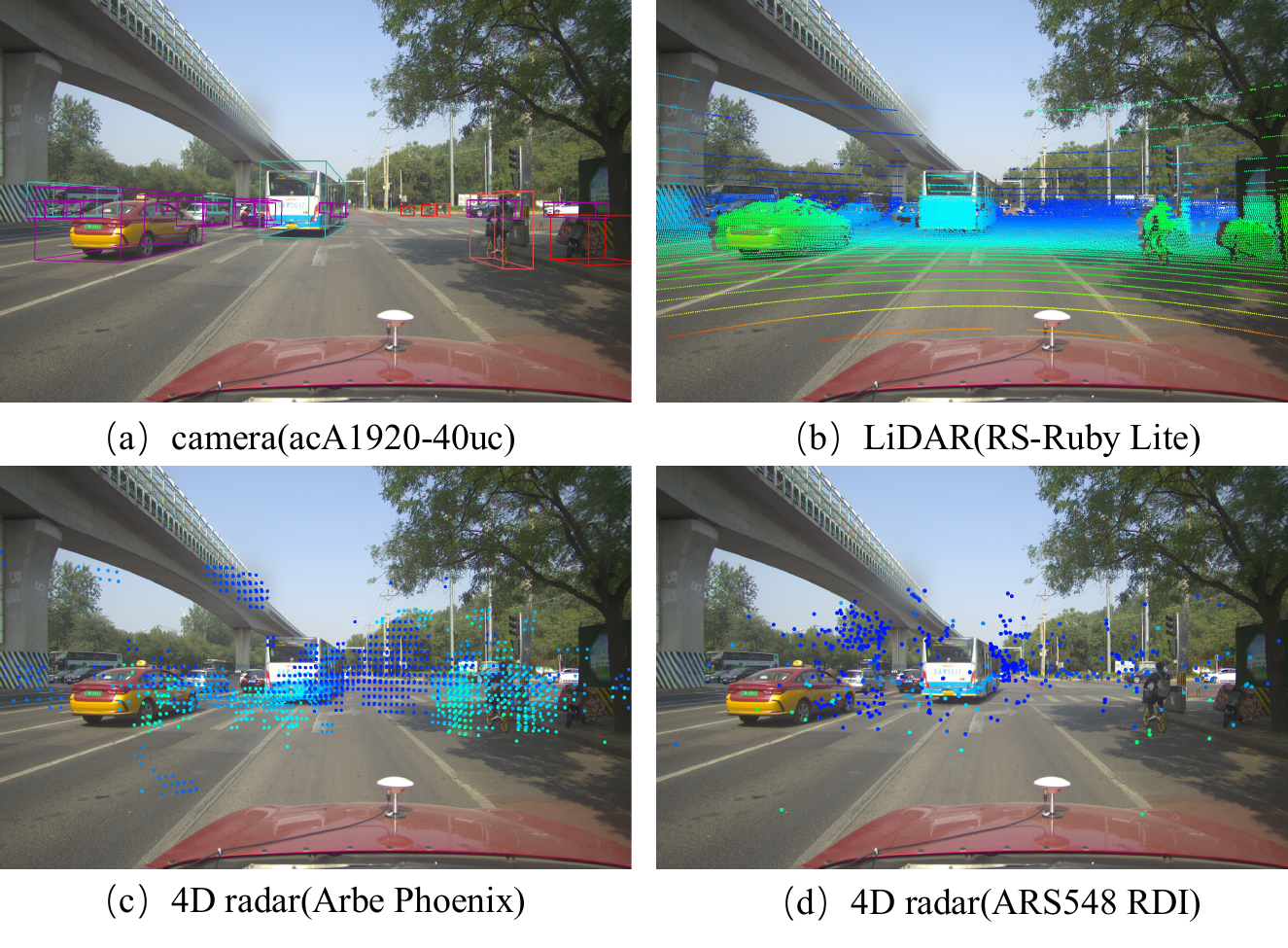}
  \caption{Projection visualization of sensor calibration. (a), (b), (c), and (d) represent the projection of the calibrated data (3D bounding box, LiDAR point cloud, Arbe Phoenix point cloud, and ARS548 RDI point cloud) on the image.  }\label{Data_calibration}
  \end{center}
\end{figure}

The calibration of this dataset is mainly categorized into joint camera-LiDAR calibration and joint camera-4D radar calibration, and they are both carried out using offline calibration. The traditional methods of calibrating multi-modal sensors have several existing works for cameras and LiDARs\cite{geiger2012automatic}. The existing 3D radar calibration methods take reference from the calibration of LiDAR and achieve good results based on the 3D radar traits\cite{el2015radar}. So, the method we obtain the internal and external parameters of 4D radar can refer to 3D radar calibration. The data calibration results are shown in Fig. \ref{Data_calibration}.

The sensors in our ego vehicle system take the origin of the LiDAR coordinate system as the origin of the multi-sensor relative coordinate system. We use offline calibration to obtain accurate calibration results easily, as well as a flexible choice of sites and calibration methods\cite{dhall2017LiDAR,pervsic2019extrinsic}. We select a plane field at normal light on a sunny day to get the calibration results of the camera. We place a rigid calibration plate at a suitable location in front of the sensor system, using a spherical coordinate system to extract and compute the external parameters of the LiDAR by the principle of realizing a rigid transformation with the 3D information of the sphere. We completed a joint camera-4D radar calibration with tools such as a corner reflector and a calibration plate, using the sensitivity of the 4D radar to recognize the signal strength of metal points to calculate and extract the calibration parameters of the 4D radar. In this case, we realized the joint calibration work for the camera, LiDAR, and 4D radar.

\begin{figure}
  \begin{center}
  \includegraphics[width=3.5in]{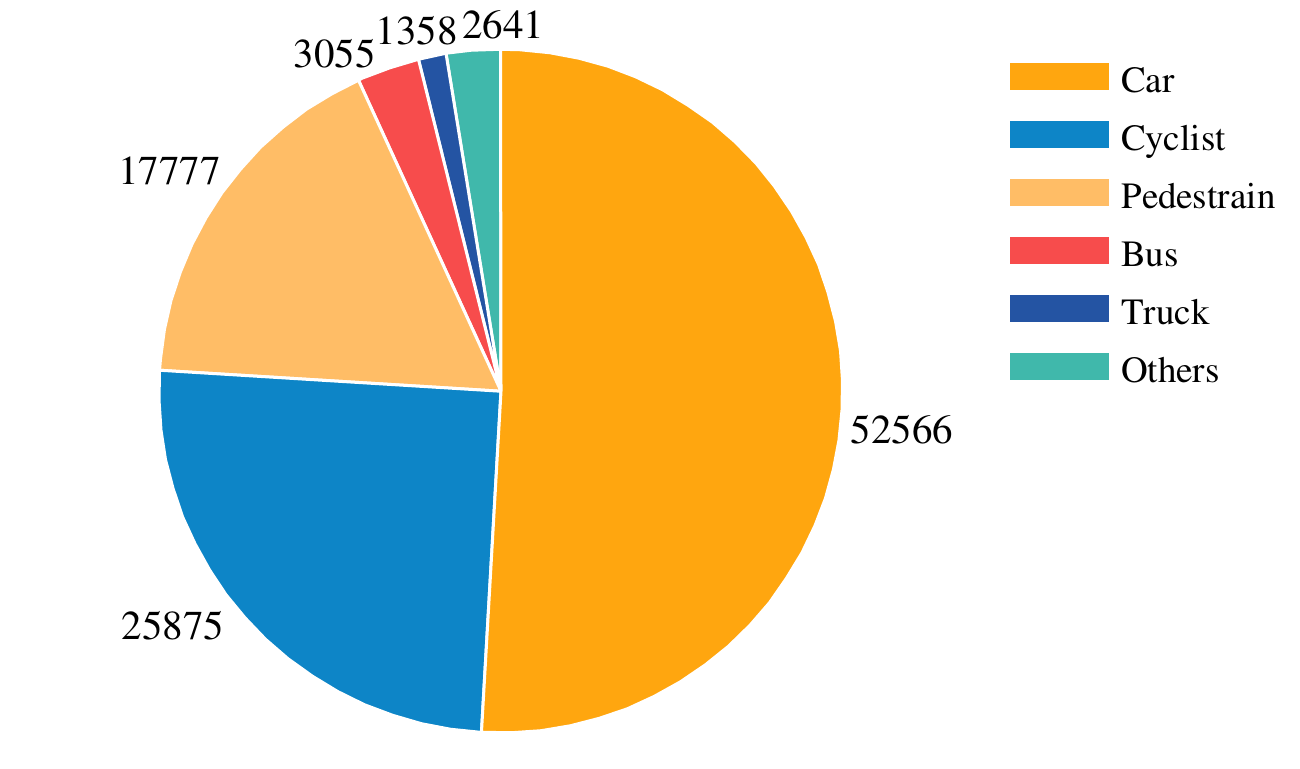}
  \caption{The statistics of the number of objects in different labels. The result suggests that the main labels like Car, Pedestrian, and Cyclist take up over three-quarters of the total amount of objects.}\label{Sector_Number_of_Objects_for_All_Categories}
  \end{center}
\end{figure}


\begin{figure}
  \begin{center}
  \includegraphics[width=3.3in]{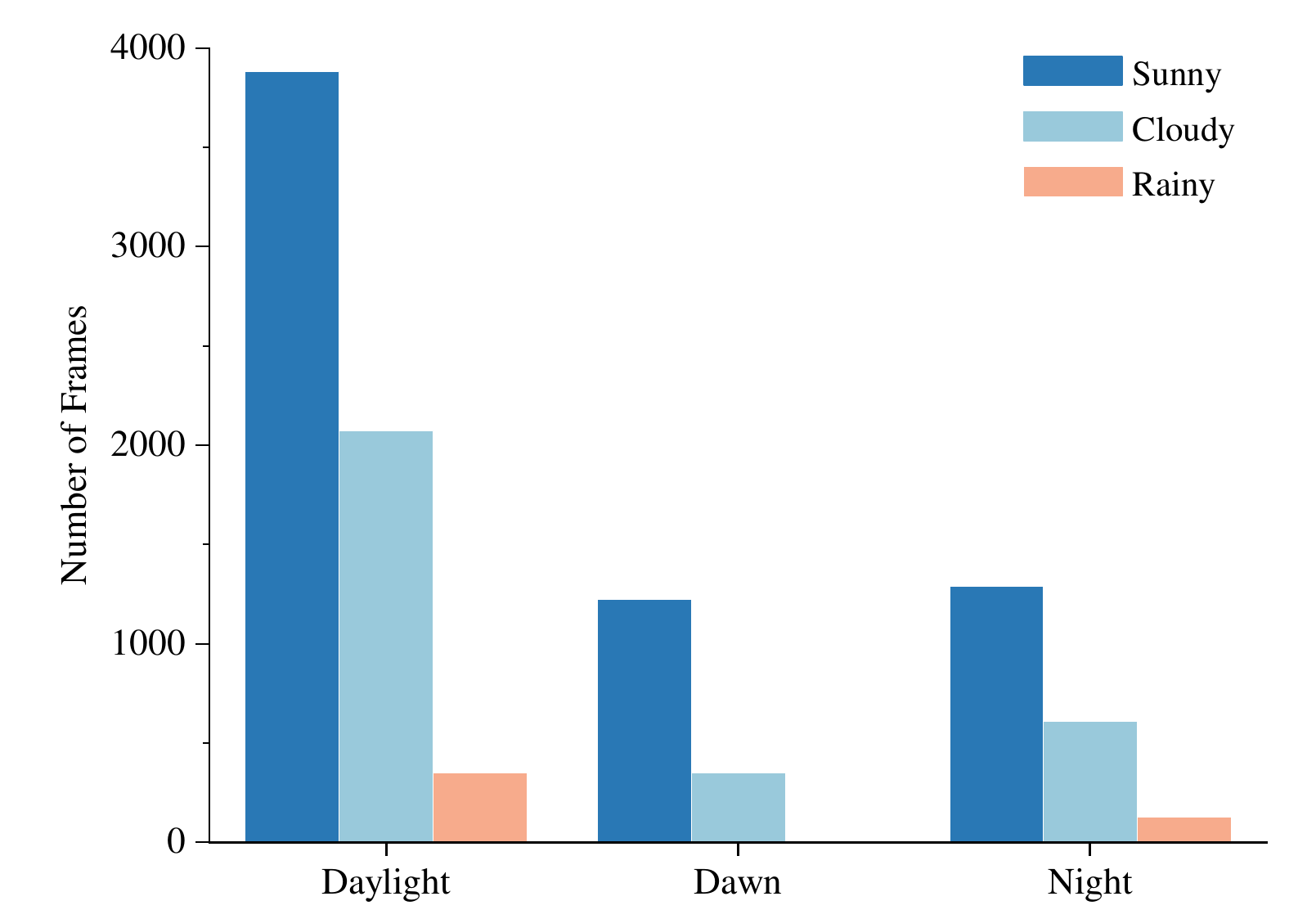}
  \caption{The statistics of the number of frames for various periods in different weather. Our dataset is classified into eight categories based on the weather conditions and periods.}\label{Weather_Number_of_Frames_in_Different_Weathers}
  \end{center}
\end{figure}
\begin{figure*}[h!]
  \begin{center}
  \includegraphics[width=6in]{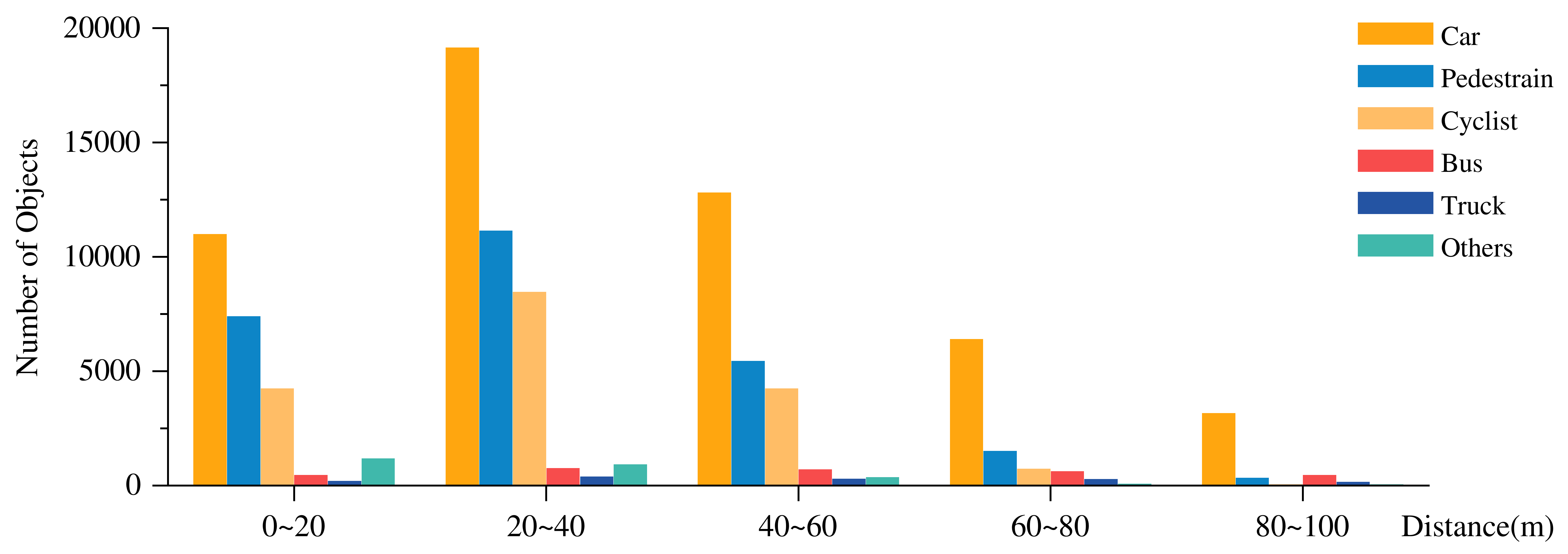}
  \caption{The statistic of different annotated objects at different ranges of distance from the ego vehicle. From the results, the majority of the annotated objects are in the range of 20m-60m. About 10$\%$ of the annotated objects are in the range of more than 60m.}\label{Distance_Number_of_Objects_in_Every_Category_from_Different_Range_of_Distance}
  \end{center}
\end{figure*} 

\begin{figure*}[h!]
  \begin{center}
  \includegraphics[width=7in]{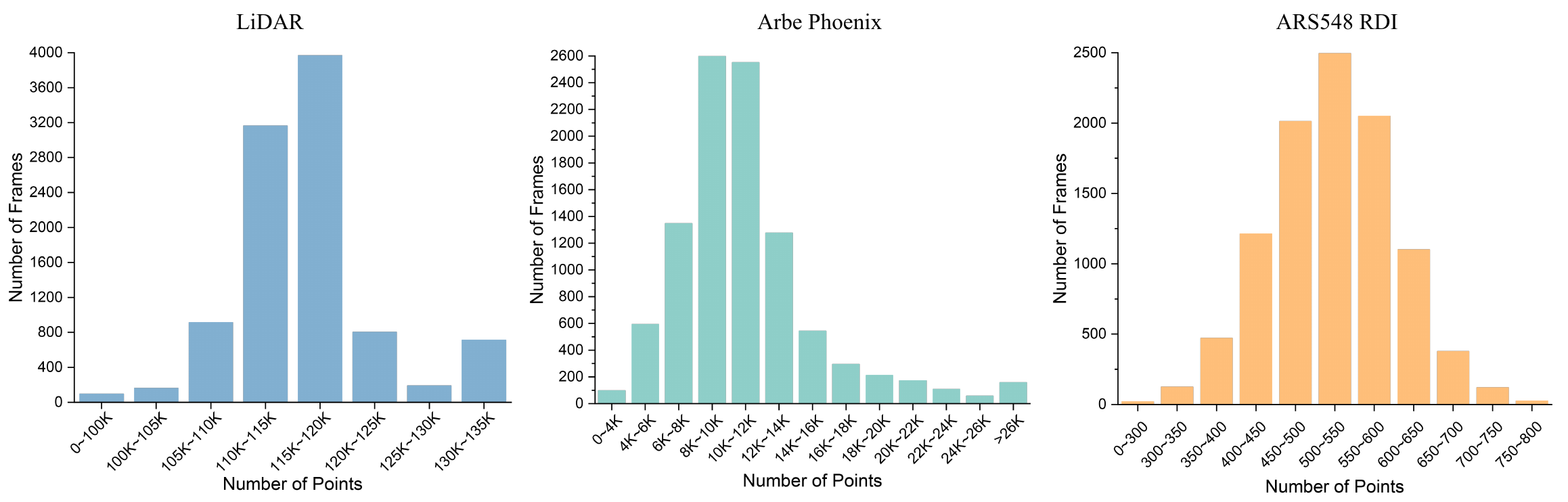}
  \caption{The statistics on the number of point clouds per data frame. The results show that most LiDAR point cloud counts are between 110K and 120K. Most Arbe Phoenix point cloud counts are between 6K and 14K. Most ARS548 RDI point cloud counts are between 400 and 650.}\label{Statistics_of_the_Number_of_Points}
  \end{center}
\end{figure*} 
\subsection { Dataset Annotation}
In our dataset, we provide the 3D bounding box, object labels, and tracking IDs for each object. We annotate the synchronized frames in time based on camera and LiDAR point clouds. The 3D bounding box we provide for each object is derived from projecting the LiDAR point cloud onto the images. We do not distinguish between objects in dynamic or static states during annotation. To synchronize the timestamps between different sensors, we choose the Precision Time Protocol (PTP) to align the time between multiple sensors using GPS message timing with time synchronization devices.

We provide basic information for each object within a 100-meter distance, including the relative coordinates (\emph{x},  \emph{y}, \emph{z}) of the corresponding 3D bounding box, absolute dimensions (length, width, height), and the orientation angle (alpha) in the Bird's Eye View (BEV) that needs careful calculation. To ensure the usability of the data, we also provide features for object occlusion and truncation. We annotated over ten categories of labels and focused on five labeling categories, including “Car”, “Pedestrian”, “Cyclist”, “Bus” and “Truck”. The remaining labels are grouped as “Others”. According to statistics from the collected raw data, approximately 50,000 synchronized frames were extracted, and 10,007 frames were annotated among them. We annotated 103,272 objects from the annotated frames.

\begin{table}[h!]
  \begin{center}
  \setlength{\tabcolsep}{5pt}
  \renewcommand\arraystretch{1.1}
  \centering
  \caption{The statistics of the number of point clouds per frame}

      \begin{tabular}{cccc}
          \toprule[1pt]
            {\textbf{Transducers}}    & {\textbf{Minimum Value}}   & {\textbf{Average Value}}   & {\textbf{Maximum Values}}   \\
          \midrule[0.4pt]
          
            LiDAR  & 74,386 & 116,096 & 133,538 \\
            Arbe Phoenix   & 898    & 11,172 & 93,721 \\
            ARS548 RDI & 243    & 523    & 800  \\
          \bottomrule[1pt]
      \end{tabular}    
      
    \label{number_of_point}
\end{center}
\end{table}

\subsection {Data Collection and Distribution}

\begin{figure*}
  \begin{center}
  \includegraphics[width=6.9in]{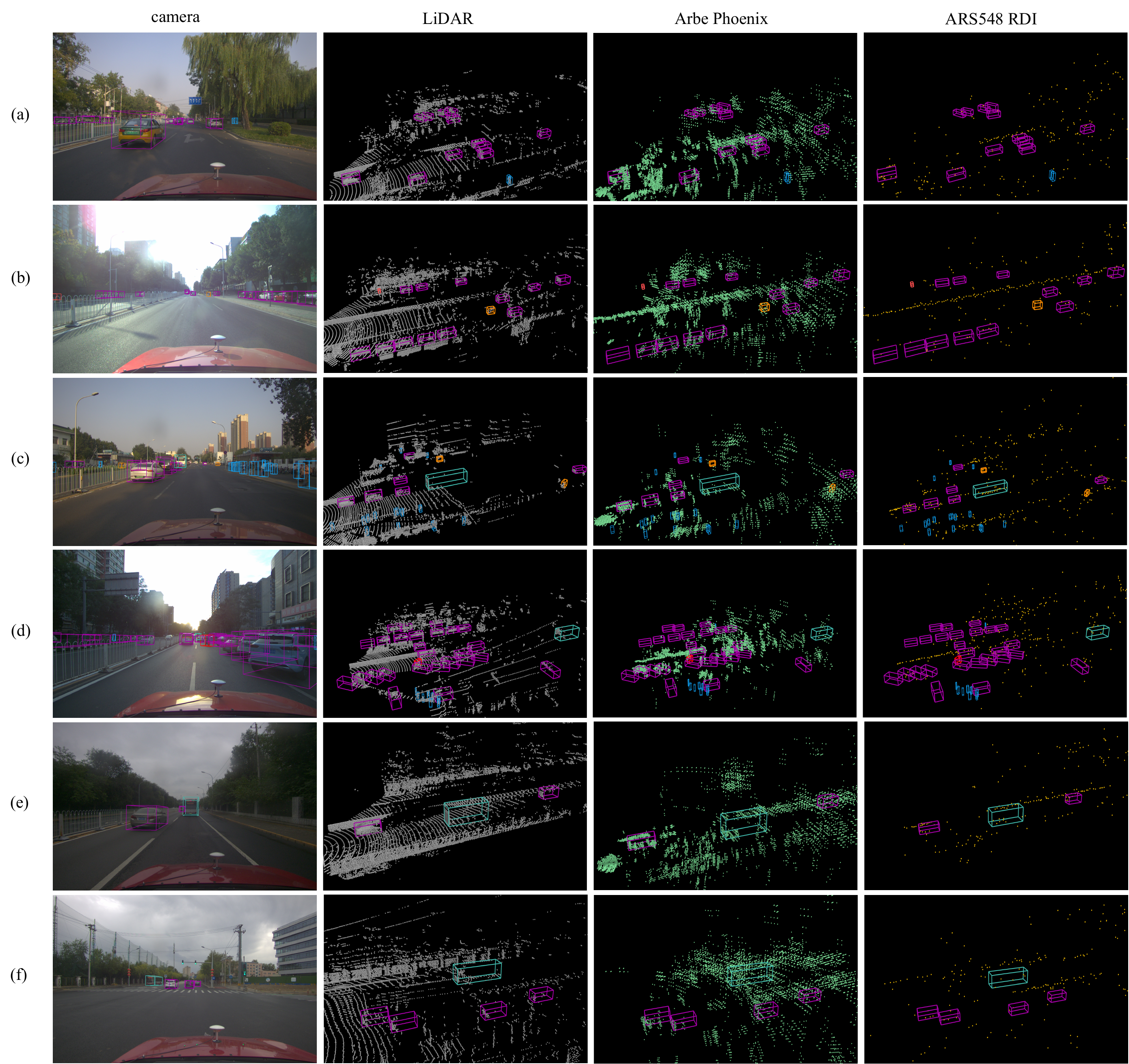}
  \caption{Representing 3D annotations in multiple scenarios and sensor modalities. The four columns respectively display the projection of 3D annotation boxes in images, LiDAR point clouds, Arbe Phoenix and ARS548 RDI radar point clouds. Each row represents a scenario type. (a) downtown daytime normal light; (b) downtown daytime backlight; (c) downtown dusk normal light; (d) downtown dusk backlight; (e) downtown clear night; (f) downtown daytime cloudy.}\label{Data_Visualization1}
  \end{center}
\end{figure*}


\begin{figure*}
  \begin{center}
  \includegraphics[width=6.9in]{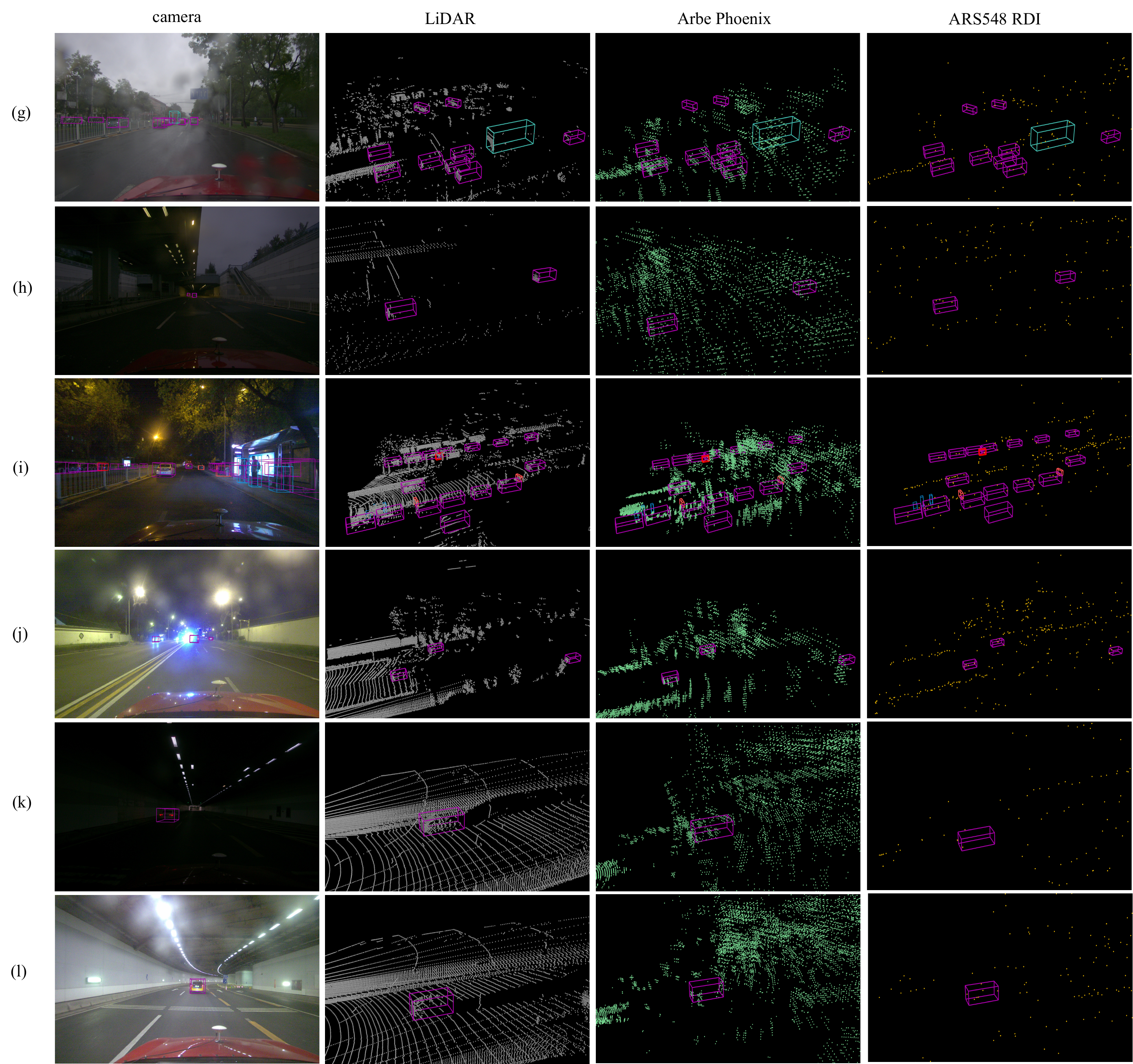}
  \caption{Representing 3D annotations in multiple scenarios and sensor modalities. Each row represents one scenario. (g) downtown rainy day; (h) downtown cloudy dusk; (i) downtown cloudy night; (j) downtown rainy night; (k) daytime tunnel; (l) nighttime tunnel.}\label{Data_Visualization2}
  \end{center}
\end{figure*}
We conducted statistical analysis on our dataset and summarized the total counts for each label, as shown in Fig. \ref{Sector_Number_of_Objects_for_All_Categories}. We presented a pie chart to display the object counts of the top six labels. The “Car” label slightly exceeds 50$\%$ of the total object count. Most of our data was obtained in urban conditions with decent traffic roads. As a result, the majority of labels are concentrated in “Car”, “Cyclist” and “Pedestrian”. We choose objects from these three labels to validate the performance of our dataset. We further analyzed the frame counts under different weather conditions and periods, as shown in Fig. \ref{Weather_Number_of_Frames_in_Different_Weathers}. About two-thirds of our data were collected under normal weather conditions, and about one-third were collected under rainy and cloudy conditions. We collected 577 frames in rainy weather, which is about 5.5$\%$ of the total dataset. The rainy weather data we collect can be used to test the performance of different 4D radars in adverse weather conditions. 

We also collected data at dawn and night when low light intensity challenged the camera's performance. We also conducted a statistical analysis of the number of objects with each label at different distance ranges from our vehicle, as shown in Fig. \ref{Distance_Number_of_Objects_in_Every_Category_from_Different_Range_of_Distance}. Most objects are within 60 meters of our ego vehicle. The distribution of distances between “Bus”, “Truck” and our ego vehicle is uniform across each range. In addition, we analyzed the distribution density of the point clouds and the number of point clouds per frame, as shown in Fig. \ref{Statistics_of_the_Number_of_Points} and Table \ref{number_of_point}. In most cases, the LiDAR has 110K to 120K points per frame, the Arbe Phoenix radar has 6K to 14K points per frame, and the ARS548 RDI radar has 400 to 650 points per frame. Due to a lot of noise in the data collection set in the tunnel scenario, there will be part of the data with much larger than normal point clouds, as shown in 713 frames of LiDAR data with more than 130K point clouds and 158 frames of Arbe Phoenix radar data with more than 26K point clouds. Based on these statistical results, our dataset encompasses adverse driving scenarios, making it conducive for the application of objects like “Car”, “Pedestrian” and “Cyclist” in experiments. Moreover, the simultaneous collection of 4D radar point clouds from two types of 4D radar is useful for analyzing the effects of different 4D radar point clouds on specific driving scenarios and researching perceptual algorithms that can process different 4D radar point clouds. This dataset has significant implications for applying theoretical experiments in autonomous driving.

\subsection{Data Visualization}

We visualize some of the data as shown in Fig. \ref{Data_Visualization1} and Fig. \ref{Data_Visualization2}. We annotate objects using 3D bounding boxes and map them to the image, the LiDAR point cloud, and two 4D radar point clouds. The 3D bounding box fits the object well and accurately depicts the corresponding points in the point cloud. The objects on both LiDAR point cloud and 4D radar point cloud corresponded well to the objects on the image, confirming good synchronization. The robustness of the 4D radar in different scenarios, weather, and light conditions can be observed in Fig. \ref{Data_Visualization1} and Fig. \ref{Data_Visualization2}. As shown in Fig. \ref{Data_Visualization2} (g, h, i, j, and k), the camera is highly affected by light and weather, and the camera image lacks RGB information to observe the object when the light is weak, backlit and rainy. In the corresponding scenarios, LiDAR acquires very tight spatial information, compensating for the camera's shortcomings. However, the LiDAR cannot effectively distinguish overlapping or relatively close objects in scenarios with more objects. The Arbe Phoenix radar can collect a dense point cloud, which can collect all the objects within the field of view, but it contains much noise. The point cloud collected by the ARS548 RDI radar is less dense, with leakage detection, but each point cloud accurately represents the object information. The ARS548 RDI radar can also be used to collect the information of the object in the field of view.

\begin{table*}[h!]
    \begin{center}
    \setlength{\tabcolsep}{2pt}
    \renewcommand\arraystretch{1.1}
    \centering
    \renewcommand\arraystretch{1.3}
    \caption{Experimental results of single model baseline for three categories}
    \resizebox{\textwidth}{!}{
    \begin{tabular}{ cc c ccc c ccc c ccc c ccc c ccc c ccc}
        \toprule[1pt]
        \multirow{3}{*}{{\textbf{Baselines}}} & \multirow{3}{*}{{\textbf{Data}}} & \multicolumn{7}{c}{{\textbf{Car}}} & & \multicolumn{7}{c}{{\textbf{Pedestrian}}} & &  \multicolumn{7}{c}{{\textbf{Cyclist}}}      \\
        \cline{3-5, 7-9, 11-13, 15-17, 19-21, 23-25} & & \multicolumn{3}{c}{{\textbf{3D@0.5}}} & & \multicolumn{3}{c}{{\textbf{BEV@0.5}}} & & \multicolumn{3}{c}{{\textbf{3D@0.25}}} & & \multicolumn{3}{c}{{\textbf{BEV@0.25}}} & & \multicolumn{3}{c}{{\textbf{3D@0.25}}} & & \multicolumn{3}{c}{{\textbf{BEV@0.25}}}      \\
        \cline{3-5, 7-9, 11-13, 15-17, 19-21, 23-25} & & {\textbf{Easy}} & {\textbf{Mod.}} & {\textbf{Hard}} & & {\textbf{Easy}} & {\textbf{Mod.}} & {\textbf{Hard}} & & {\textbf{Easy}} & {\textbf{Mod.}} & {\textbf{Hard}} & & {\textbf{Easy}} & {\textbf{Mod.}} & {\textbf{Hard}} & & {\textbf{Easy}} & {\textbf{Mod.}} & {\textbf{Hard}} & & {\textbf{Easy}} & {\textbf{Mod.}} & {\textbf{Hard}}\\
    \midrule[0.4pt]
\multirow{3}{*}{PointPillars\cite{lang2019pointpillars}} & LiDAR  & 81.78 & 55.40 & 44.53 &  & 81.81 & 55.49 & 45.69 &  & 43.22 & 38.87 & 38.45 &  & 43.60 & 39.59 & 38.92 &  & 25.60 & 24.35 & 23.97 &  & 38.78 & 38.74 & 38.42 \\
                              & Arbe   & 49.06 & 27.64 & 18.63 &  & 54.63 & 35.09 & 25.19 &  & 0.00  & 0.00  & 0.00  &  & 0.00  & 0.00  & 0.00  &  & 0.19  & 0.12  & 0.12  &  & 0.41  & 0.24  & 0.23  \\
                              & ARS548 & 11.94 & 6.12  & 3.76  &  & 14.40 & 8.14  & 5.26  &  & 0.00  & 0.00  & 0.00  &  & 0.01  & 0.01  & 0.01  &  & 0.99  & 0.63  & 0.58  &  & 2.27  & 1.64  & 1.53 \\
\midrule[0.4pt]
\multirow{3}{*}{Voxel R-CNN\cite{deng2021voxel}} & LiDAR  & \textbf{86.41} & 56.91 & 42.38 &  & \textbf{86.41} & 56.95 & 42.43 &  & 52.65 & 46.33 & 45.80 &  & 53.50 & 46.46 & 45.93 &  & 38.89 & 35.13 & 34.52 &  & 47.47 & 45.43 & 43.85 \\
                           & Arbe   & 55.47 & 30.17 & 19.82 &  & 59.32 & 34.86 & 23.77 &  & 0.03  & 0.02  & 0.02  &  & 0.02  & 0.02  & 0.02  &  & 0.15  & 0.06  & 0.06  &  & 0.21  & 0.15  & 0.15  \\
                           & ARS548 & 18.37 & 8.24  & 4.97  &  & 21.34 & 9.81  & 6.11  &  & 0.00  & 0.00  & 0.00  &  & 0.00  & 0.00  & 0.00  &  & 0.24  & 0.21  & 0.21  &  & 0.33  & 0.30  & 0.30  \\
\midrule[0.4pt]
\multirow{3}{*}{RDIoU\cite{sheng2022rethinking}} & LiDAR  & 63.43 & 40.80 & 32.92 &  & 63.44 & 41.25 & 33.74 &  & 33.71 & 29.35 & 28.96 &  & 33.97 & 29.62 & 29.22 &  & 38.26 & 35.62 & 35.02 &  & 49.33 & 47.48 & 46.85 \\
                       & Arbe   & 51.49 & 26.74 & 17.83 &  & 55.27 & 31.48 & 21.80 &  & 0.00  & 0.00  & 0.00  &  & 0.01  & 0.01  & 0.01  &  & 0.51 & 0.37  & 0.35  &  & 0.84  & 0.66  & 0.65  \\
                       & ARS548 & 5.96  & 3.77  & 2.29  &  & 7.13  & 5.00  & 3.21  &  & 0.00  & 0.00  & 0.00  &  & 0.00  & 0.00  & 0.00  &  & 0.21  & 0.15  & 0.15  &  & 0.61  & 0.46  & 0.44 \\
\midrule[0.4pt]

\multirow{3}{*}{CasA-V\cite{wu2022casa}} & LiDAR  & 80.60 & \textbf{58.98} & \textbf{49.83} &  & 80.60 & \textbf{59.12} & \textbf{51.17} &  & 55.43 & 49.11 & 48.47 &  & 55.66 & 49.35 & 48.72 &  & \textbf{42.84} & \textbf{40.32} & \textbf{39.09} &  & \textbf{51.51} & \textbf{50.03} & \textbf{49.35} \\
                        & Arbe   & 27.96 & 10.27 & 6.21  &  & 30.52 & 12.28 & 7.82  &  & 0.02  & 0.01  & 0.01  &  & 0.02  & 0.02  & 0.02  &  & 0.05  & 0.04  & 0.04  &  & 0.13  & 0.05  & 0.05  \\
                        & ARS548 & 7.71  & 3.05  & 1.86  &  & 8.81  & 3.74  & 2.38  &  & 0.00  & 0.00  & 0.00  &  & 0.00  & 0.00  & 0.00  &  & 0.08  & 0.06  & 0.06  &  & 0.25  & 0.21  & 0.19  \\
\midrule[0.4pt]
\multirow{3}{*}{CasA-T\cite{wu2022casa}} & LiDAR  & 73.41 & 45.74 & 35.09 &  & 73.42 & 45.79 & 35.31 &  & \textbf{58.84} & \textbf{52.08} & \textbf{51.45} &  & \textbf{59.06} & \textbf{52.36} & \textbf{51.74} &  & 35.42 & 33.78 & 33.36 &  & 44.35 & 44.41 & 42.88 \\
                        & Arbe   & 14.15 & 6.38  & 4.27  &  & 22.9  & 13.06 & 9.18  &  & 0.00  & 0.00  & 0.00  &  & 0.00  & 0.00  & 0.00  &  & 0.09  & 0.06  & 0.05  &  & 0.17  & 0.08  & 0.08  \\
                        & ARS548 & 3.16  & 1.60  & 1.00  &  & 4.21  & 2.21  & 1.49  &  & 0.00  & 0.00  & 0.00  &  & 0.00  & 0.00  & 0.00  &  & 0.36  & 0.20  & 0.20  &  & 0.68  & 0.43  & 0.42 \\
        \bottomrule[1pt]
    \end{tabular}
    }
    \begin{tablenotes}{    
          \footnotesize   
          \item[1] “Arbe”: “Arbe Phoenix”. “ARS548”: “ARS548 RDI”. “Mod.”: “Moderate”. 
          }
    \end{tablenotes}   
        
  \label{tab:experiment_iou0.5_single}
\end{center}
\end{table*}

\begin{table*}[h!]
    \begin{center}
    \setlength{\tabcolsep}{2pt}
    \renewcommand\arraystretch{1.1}
    \centering
    \renewcommand\arraystretch{1.3}
    \caption{ Experimental results of mutiple modal baseline for three categories }
    \resizebox{\textwidth}{!}{
    \begin{tabular}{ cc c ccc c ccc c ccc c ccc c ccc c ccc}
        \toprule[1pt]
        \multirow{3}{*}{{\textbf{Baselines}}} & \multirow{3}{*}{{\textbf{Data}}} & \multicolumn{7}{c}{{\textbf{Car}}} & & \multicolumn{7}{c}{{\textbf{Pedestrian}}} & &  \multicolumn{7}{c}{{\textbf{Cyclist}}}      \\
        \cline{3-5, 7-9, 11-13, 15-17, 19-21, 23-25} & & \multicolumn{3}{c}{{\textbf{3D@0.5}}} & & \multicolumn{3}{c}{{\textbf{BEV@0.5}}} & & \multicolumn{3}{c}{{\textbf{3D@0.25}}} & & \multicolumn{3}{c}{{\textbf{BEV@0.25}}} & & \multicolumn{3}{c}{{\textbf{3D@0.25}}} & & \multicolumn{3}{c}{{\textbf{BEV@0.25}}}      \\
        \cline{3-5, 7-9, 11-13, 15-17, 19-21, 23-25} & & {\textbf{Easy}} & {\textbf{Mod.}} & {\textbf{Hard}} & & {\textbf{Easy}} & {\textbf{Mod.}} & {\textbf{Hard}} & & {\textbf{Easy}} & {\textbf{Mod.}} & {\textbf{Hard}} & & {\textbf{Easy}} & {\textbf{Mod.}} & {\textbf{Hard}} & & {\textbf{Easy}} & {\textbf{Mod.}} & {\textbf{Hard}} & & {\textbf{Easy}} & {\textbf{Mod.}} & {\textbf{Hard}}\\
    \midrule[0.4pt]

\multirow{3}{*}{\makecell{VFF \\ \cite{li2022voxel}} } & Carmera+LiDAR & \textbf{94.60} & \textbf{84.14} & \textbf{78.77} &  & \textbf{94.60} & 84.28 & \textbf{80.55} &  & \textbf{39.79} & \textbf{35.99} & \textbf{36.54} &  & \textbf{40.32} & \textbf{36.59} & \textbf{37.28} &  & \textbf{55.87} & \textbf{51.55} & \textbf{51.00} &  & \textbf{55.87} & \textbf{51.55} & \textbf{51.00} \\
                     & Carmera+Arbe  & 31.83 & 14.43 & 11.30 &  & 36.09 & 17.20 & 13.23 &  & 0.01  & 0.01  & 0.01  &  & 0.01  & 0.01  & 0.01  &  & 0.20  & 0.07  & 0.08  &  & 0.20  & 0.08  & 0.08 \\
                              & Carmera+ARS548  &   12.60   &    6.53     &  4.51 &  & 16.34 &  9.58 &   6.61   &  &   0.00  &  0.00 &  0.00   &  &  0.00 &  0.00  & 0.00  &  &  0.00   &   0.00  &  0.00    &  &  0.00 &  0.00  &  0.00                      \\
\midrule[0.4pt]
\multirow{2}{*}{\makecell{M$^2$-Fusion \\ \cite{wang2022multi}} } &LiDAR+Arbe & 89.71 & 79.70 & 64.32 &  & 90.91 & \textbf{85.73} & 70.16 &  & 27.79 & 20.41 & 19.58 &  & 28.05 & 20.68 & 20.47 &  & 41.85 & 36.20 & 35.14 &  & 42.60 & 36.79 & 36.03 \\
                          & LiDAR+ARS548     & 89.91 & 78.17 & 62.37 &  & 91.14 & 82.57 & 66.65 &  & 34.28 & 29.89 & 29.17 &  & 34.98 & 30.28 & 29.92 &  & 42.42 & 40.92 & 39.98 &  & 43.12 & 41.57 & 40.29\\
                              
        \bottomrule[1pt]
    \end{tabular}
    } 
        
  \label{tab:experiment_iou0.5_mul}
\end{center}
\end{table*}

\section{Experiment}\label{sec4:experiment}

This section establishes the experimental platform for conducting the experiments on our dataset. We utilize several state-of-the-art baselines to validate our dataset. The performance of our dataset is verified by the results obtained from the experiments. We then conduct both qualitative and quantitative analyses on the results of the experiments, and finally, we get our evaluations of our dataset.

\subsection{Experiment Settings}

We employed a server based on the Ubuntu 18.04 system as our hardware platform. We adopted the OpenPCDet project based on Pytorch 10.2, the batch size setting to the default value of 4, and each experiment was trained for 80 epochs at a learning rate of 0.003 on four Nvidia RTX3090 graphics cards. To demonstrate the performance of our dataset and the designed algorithms, we introduced some existing state-of-the-art algorithms, all carrying out experiments as required.

\begin{table*}[h!]
    \begin{center}
    \setlength{\tabcolsep}{2pt}
    \renewcommand\arraystretch{1.1}
    \centering
    \renewcommand\arraystretch{1.3}
    \caption{ Experimental results for three categories in the rainy scenario }
    \resizebox{\textwidth}{!}{
    \begin{tabular}{ cc c ccc c ccc c ccc c ccc c ccc c ccc}
        \toprule[1pt]
        \multirow{3}{*}{{\textbf{Baselines}}} & \multirow{3}{*}{{\textbf{Data}}} & \multicolumn{7}{c}{{\textbf{Car}}} & & \multicolumn{7}{c}{{\textbf{Pedestrian}}} & &  \multicolumn{7}{c}{{\textbf{Cyclist}}}      \\
        \cline{3-5, 7-9, 11-13, 15-17, 19-21, 23-25} & & \multicolumn{3}{c}{{\textbf{3D@0.5}}} & & \multicolumn{3}{c}{{\textbf{BEV@0.5}}} & & \multicolumn{3}{c}{{\textbf{3D@0.25}}} & & \multicolumn{3}{c}{{\textbf{BEV@0.25}}} & & \multicolumn{3}{c}{{\textbf{3D@0.25}}} & & \multicolumn{3}{c}{{\textbf{BEV@0.25}}}      \\
        \cline{3-5, 7-9, 11-13, 15-17, 19-21, 23-25} & & {\textbf{Easy}} & {\textbf{Mod.}} & {\textbf{Hard}} & & {\textbf{Easy}} & {\textbf{Mod.}} & {\textbf{Hard}} & & {\textbf{Easy}} & {\textbf{Mod.}} & {\textbf{Hard}} & & {\textbf{Easy}} & {\textbf{Mod.}} & {\textbf{Hard}} & & {\textbf{Easy}} & {\textbf{Mod.}} & {\textbf{Hard}} & & {\textbf{Easy}} & {\textbf{Mod.}} & {\textbf{Hard}}\\
    \midrule[0.4pt]
\multirow{3}{*}{ PointPillars\cite{lang2019pointpillars} } & LiDAR  & 60.57 & 44.31 & 41.91 &  & 60.57 & 44.56 & 42.49 &  & \textbf{32.74} & \textbf{28.82} & \textbf{28.67} &  & \textbf{32.74} & \textbf{28.82} & \textbf{28.67} &  & 29.12  & 25.75 & 24.24 &  & 44.39  & 40.36 & 38.64 \\
                              & Arbe  & \textbf{68.24} & 48.98 & 42.80 &  & \textbf{74.50} & \textbf{59.68} & \textbf{54.34} &  & 0.00  & 0.00  & 0.00  &  & 0.00  & 0.00  & 0.00  &  & 0.19   & 0.10  & 0.09  &  & 0.32   & 0.16  & 0.15  \\
                              & ARS548 & 11.87  & 8.41  & 7.32  &  & 14.16  & 11.32  & 9.82  &  & 0.11  & 0.09  & 0.08  &  & 0.11  & 0.09  & 0.08  &  & 0.93   & 0.36  & 0.30  &  & 2.26   & 1.43  & 1.20  \\
\midrule[0.4pt]
\multirow{3}{*}{ RDIoU\cite{sheng2022rethinking} } & LiDAR  & 44.93 & 39.32 & 39.09 &  & 44.93 & 39.39 & 39.86 &  & 24.28 & 21.63 & 21.43 &  & 24.28 & 21.63 & 21.43 &  & \textbf{52.64} & \textbf{43.92} & \textbf{42.04} &  & \textbf{60.80} & \textbf{52.44} & \textbf{50.28} \\
                       & Arbe  & 67.81 & \textbf{49.59} & \textbf{43.24} &  & 70.09 & 54.17 & 47.64 &  & 0.00  & 0.00  & 0.00  &  & 0.00  & 0.00  & 0.00  &  & 0.38  & 0.30  & 0.28  &  & 0.63  & 0.45  & 0.45  \\
                       & ARS548 & 5.87  & 5.48  & 4.68  &  & 6.36  & 6.51 & 5.46  &  & 0.00  & 0.00  & 0.00  &  & 0.00  & 0.00  & 0.00  &  & 0.09  & 0.01  & 0.01  &  & 0.13  & 0.08  & 0.08\\

        \bottomrule[1pt]
    \end{tabular}
    } 
        
  \label{tab:experiment_iou0.5_rain}
\end{center}
\end{table*}

\subsection{Experiment Details}

We take three kinds of labels as the experiment objects, which are “Car”, “Pedestrian” and “Cyclist”. According to the need of the experiment, we split all the annotated data into a training, validation, and test set according to the ratio of 51$\%$, 25$\%$, and 24$\%$. Our dataset will be published in the KITTI format. We use Average Precision (AP) as a performance measure. For AP, we set overlap for different labels considering the IoU and the effects, such as 3D bounding box bias due to ground truth in various scenarios. We set the overlap at 50$\%$ for the “Car” category and 25$\%$ for the “Pedestrian” and “Cyclist” category. To evaluate our dataset, we conduct single-modal experiments and multi-modal experiments with several state-of-the-art baseline models, and the results are shown in Table \ref{tab:experiment_iou0.5_single} and Table \ref{tab:experiment_iou0.5_mul}.

\subsection{Quantitative Analysis}

Since the baseline model performs the same for all three levels of occlusion for representational convenience, this section only discusses the case when the object occlusion level is moderate.

As can be seen in Table \ref{tab:experiment_iou0.5_single}, the LiDAR point clouds captured in our dataset have excellent detection results, while the 4D radar point clouds have the potential for improvement. In the BEV view, the detection accuracies of the CasA-V model using the LiDAR point cloud reach 59.12$\%$ (Car), 49.35$\%$ (Pedestrian), and 50.03$\%$ (Cyclist), which indicates that the LiDAR point cloud in this dataset is able to represent the object information well. However, the detection accuracies are lower when using the 4D radar. In the BEV view, it can be observed that the CasA-V model achieves 12.28$\%$ and 3.74$\%$ in the “Car” category using the Arbe Phoenix and ARS548 RDI point clouds. This is because the 80-line LiDAR has a higher point density and collects much more information than the 4D radar. Meanwhile, the PointPillars model has achieved an accuracy of 35.09$\%$ in the category of “Car” when using the Arbe Phoenix radar point cloud, which is 26.95$\%$ higher than ARS548 RDI. This shows that despite the noise, the Arbe Phoenix radar point cloud can still represent the object information. The ARS548 RDI radar point cloud outperforms the Arbe Phoenix radar point cloud in the middle-sized “Cyclist” category. The accuracy of the PointPillars model on the ARS548 RDI radar point cloud reached 1.64$\%$ (Cyclist), which is 1.4$\%$ higher than the Arbe Phoenix radar point cloud. This is because cars with a large amount of metal are more conducive to collecting object information by LiDAR and radar. Although the Arbe Phoenix collects much noise, it can still better detect large metal objects. While the “Cyclist” object has less metal on it, collecting a lot of noise affects the model's performance. The dataset label data defines the 3D bounding box through LiDAR point clouds, which cannot sometimes be captured by 4D radar because “pedestrian” objects usually have a smaller area, and 4D radar point clouds are sparser than LiDAR point clouds. When using the 4D radar point clouds for training, there is a situation where there is no object in the label that causes the final result to be poor or even zero.

In addition, we discuss the detection of the multi-modal baseline model using this dataset. As can be seen in Table \ref{tab:experiment_iou0.5_mul}, when the M$^2$-Fusion model fuses LiDAR and 4D radar, the detection of the “Car” category has a huge performance improvement. The detection results in the BEV view are higher than those of the other single-modal baseline models and 30.24$\%$ higher than the model of PointPillars. This indicates that the 4D radar data of this dataset can provide auxiliary information for LiDAR. Meanwhile, in the 3D view of the “Cyclist” category, when the LiDAR and ARS548 RDI radar point clouds are fused, the result is 4.72$\%$ higher than the result of fusing LiDAR and Arbe Phoenix radar point clouds, and 0.6$\%$ higher than the CasA-V model using only LiDAR point clouds. This further demonstrates the superior detection performance of the ARS548 RDI radar point cloud for the “Cyclist” category. Moreover, the experimental results of the VFF model using the camera image fused with the LiDAR point cloud are much higher than most single-modal baseline models. In the BEV view, the VFF model using images and LiDAR has 25.16$\%$ higher detection results than the CasA-V model using only LiDAR in the “Car” category, which suggests that the camera can provide rich information and can perform well after fusing with the spatial information of the point cloud.

Finally, we compared the PointPillars and RDIoU models' detection performance in rainy scenarios. As can be seen in Table \ref{tab:experiment_iou0.5_rain}, the 4D radar performs better in rainy scenarios. For the “Car” category, the RDIoU model has much higher detection results on radar point clouds than on LiDAR point clouds. The RDIoU model's BEV and 3D view detection performance using the Arbe Phoenix radar point cloud is 14.78$\%$ and 10.27$\%$ better than LiDAR's. 
However, for the “Cyclist” and “Pedestrian” categories, both 4D radar point clouds did not achieve satisfactory results. 
The PointPillars model achieved excellent results on the ARS548 RDI point cloud, while the RDIoU achieved excellent results on the Arbe Phoenix point cloud gave better results. These experimental results show that this dataset's two 4D radar point clouds show different performance characteristics. This provides data support for further research on using 4D radar point clouds with different point cloud densities and noise levels in adverse scenarios and contributes to developing algorithms that can perform well on different 4D radar point clouds.

\section{Conclusion and future work}\label{sec5:conclusion}


We propose a large-scale multi-modal dataset with two different types of 4D radar available for both 3D object detection and tracking tasks in autonomous driving. We collect data frames in different scenarios and weather, which is useful for evaluating the performance of different 4D radars in different scenarios. It also helps to study the sensing algorithms that can process different 4D radar point clouds. We verify with the latest baseline that our dataset aligns with our expected needs. Our dataset is competent for the current perception tasks for autonomous driving.

The data we collected for various adverse weather issues did not meet expectations. In the future, we will collect a wider range of scenarios, including roads and many adverse weather conditions. These data can scale up the dataset and enhance the generalization capability of the dataset for object detection and tracking tasks. Especially in complex scenarios, the performance of 4D radar can be effectively verified. To perform well in multiple scenarios, we will continue to complement the dataset's properties, particularly in rainy, snowy, and foggy weather conditions.

\bibliographystyle{IEEEtran}
\bibliography{IEEEabrv,Bibliography.bib}

\newpage

\vfill

\end{document}